# The mechanism underlying successful deep learning

**Yarden Tzach[1], Yuval Meir[1], Ofek Tevet[1], Ronit D. Gross[1], Shiri Hodassman[1], Roni Vardi[2] and Ido Kanter[1,2*]**

[1]Department of Physics, Bar-Ilan University, Ramat-Gan, 52900, Israel.
[2]Gonda Interdisciplinary Brain Research Center, Bar-Ilan University, Ramat-Gan, 52900, Israel.

[*]Corresponding author email: ido.kanter@biu.ac.il

Deep architectures consist of tens or hundreds of convolutional layers (CLs) that terminate with a few fully connected (FC) layers and an output layer representing the possible labels of a complex classification task. According to the existing deep learning (DL) rationale, the first CL reveals localized features from the raw data, whereas the subsequent layers progressively extract higher-level features required for refined classification. This article presents an efficient three-phase procedure for quantifying the mechanism underlying successful DL. First, a deep architecture is trained to maximize the success rate (SR). Next, the weights of the first several CLs are fixed and only the concatenated new FC layer connected to the output is trained, resulting in SRs that progress with the layers. Finally, the trained FC weights are silenced, except for those emerging from a single filter, enabling the quantification of the functionality of this filter using a correlation matrix between input labels and averaged output fields, hence a well-defined set of quantifiable features is obtained. Each filter essentially selects a single output label independent of the input label, which seems to prevent high SRs; however, it counterintuitively identifies a small subset of possible output labels. This feature is an essential part of the underlying DL mechanism and is progressively sharpened with layers, resulting in enhanced signal-to-noise ratios and SRs. Quantitatively, this mechanism is exemplified by the VGG-16, VGG-6, and AVGG-16. The proposed mechanism underlying DL provides an accurate tool for identifying each filter's quality and is expected to direct additional procedures to improve the SR, computational complexity, and latency of DL.

## Introduction

The earliest artificial adaptive classifier, the Perceptron[1,2], was introduced approximately seven decades ago and consists of an input layer and a single Boolean output unit. Its learning algorithm has been inspired by brain dynamics, in which synaptic plasticity modifies the connection strength between two neurons in response to their relative activities[3,4]. Although the Perceptron learning algorithm converges in the case of existing solutions, it can only implement linearly separable classification tasks because its architecture does not consist of hidden layers. Using statistical mechanical methods for disordered systems[5], the classification capabilities of limited architectures with one hidden layer have been theoretically estimated[6-11], but without converging learning algorithms[12]. Hence, the one-hidden-layer architecture classification capabilities achieved theoretically outperform the existing algorithmic procedures, however its theoretical extension to more than one hidden-layer architecture is implausible.

The performance of complex and practical classification tasks requires more structured feedforward architectures with numerous convolutional and fully connected hidden layers, which can number in the hundreds[13,14]. The training of this type of deep architectures requires nonlocal training techniques such as backpropagation (BP)[15-17], which can guarantee convergence to local minima only. These are the two essential components of the current implementation of deep learning (DL) algorithms. Their underlying rationality is that the first convolutional layer is sensitive to the appearance of a given pattern or symmetry in limited areas of the input, whereas large-scale features characterizing a class of inputs are progressively revealed in the subsequent convolutional layers[18-21]. This hand-wave argument does not suggest a well-defined set of features, nor a quantitative way to measure their progressive appearance with increasing layers.

The current status of DL is that practical solutions for complex classification tasks precede much of our knowledge of their underlying mechanism[19,22-25]. Typically, for a given classification task and database, deeper architectures with enhanced success rates (SRs) are eventually presented, but without much quantitative theoretical understanding of the mechanism underlying learning.

To address this shortcoming, this work presents the mechanism underlying DL, which enables to quantify the progress of SRs with increasing layers and the functionality of each filter in a layer. Each filter essentially identifies a small subset of possible output labels, a feature that is progressively sharpened with layers, resulting in an enhanced signal-to-noise ratio and SR. The formation of a theoretical framework for DL is expected to enable a quantitative comparison among different deep architectures, as well as to direct advanced procedures to improve the SRs of such architectures while reducing their computational complexity and latencies.

The methodology and results are first presented in detail for VGG-16[22] and later extended and compared to those of VGG-6[22] and AVGG-16[26].

**RESULTS**

**SRs of progressive layers**

For a given trained deep architecture, the SR of each layer was quantified using the presented methodology, as first exemplified by VGG-16, which was trained to classify the CIFAR-10 database[18] (Fig. 1A). The architecture consists of 16 layers[22]; 13 convolutional layers (CLs), including 5 max-pooling (MP) operators and 3 fully connected (FC) layers, and terminates with an output layer representing the 10 possible output labels.

The SR of the $m^{th}$ layer after a max-pooling (MP) operation ($m = 2, 4, 7, 10 \ 13$), was obtained using the following procedure, without affecting the weights of the lower layers ($< m$). The $N(m)$ outputs of the $m^{th}$ layer were FC to the 10 output units, for example, $N(13) = 512$ and $N(2) = 16{,}384$ (Fig. 1B-C). These $N(m)$ outputs represented the preprocessing of the inputs by the first $m$ CLs[27,28]. Next, the $10 \cdot N(m)$ weights (Fig. 1B-C, orange) were trained to maximize the SRs using the optimized BP parameters (Supplementary Information), whereas the lower level weights of the CLs remained fixed. Optimizing a single FC layer was much simpler than optimizing the entire deep architecture. The results indicate that the average SRs increase with $m$ and exhibit small fluctuations among the samples (Fig. 2). In addition, the average SRs from at least $m \geq 10$ are practically the same as those for VGG-16. Hence, 11 layers could produce maximal SRs, while training was accomplished using 16 layers. This observation implies an efficient procedure for minimizing test latency that is independent of the recently suggested layer-folding mechanism[29]. The presented methodology for measuring the progressive layer SRs (Fig. 2) serves as a quantitative component for understanding the mechanism underlying DL.

**Single-filter SR**

The trained FC layer that maximized the $m^{th}$ layer SRs (Figs. 1-2) enabled the measurement of the SR of a single filter. First, all FC weights were silenced except for those emerging from a single filter (Fig. 3A). Next, each test input was preprocessed using the first $m-1$ layers and the single active filter of the $m^{th}$ layer, which finally generated 10 output fields using the active weights of the FC layer (Fig. 3A, brown weights). The output unit with the maximum field was selected as the predicted label (Supplementary Information).

A single-filter mutual probability between the input and output labels was summarized by a $10 \times 10$ matrix, where element $(i,j)$ represents the probability of an output label $j$ generated by $1,000$ test inputs of label $i$ (Fig. 3B). Each matrix row was normalized to unity, and inputs with zero fields on all output units were excluded. The results indicate that in layer 13 each one of the filters selects a single output label, independent of the input label (Fig. 3B, left). This feature prevents high SRs, because there is no mutual information between the input and output labels. A possible means of overcoming this essential limitation requires that for the filter selecting label $j$ independent of the input label, input $j$ generates a dominant output field. However, it is not the case for a non-negligible fraction of inputs. A similar preference also characterizes most of the filters in the lower layers ($m <$ 13), where one or a few output labels are selected independently of the input label (Fig. 3B, right). Hence, the mechanism underlying DL appears to be unassociated with single-filter SRs.

**Single-filter performance**

The single-filter performance was calculated where all weights of the concatenated FC layer to the $m^{th}$ layer were silenced, except for those emerging from a single filter (Fig. 3A). The results were depicted by a $10 \times 10$ color-coded matrix, where element $(i,j)$ represents the sum of $1,000$ fields generated by label $i$ test inputs on output $j$. Finally, the matrix elements were normalized by their maximal element (Fig. 4, left column). Three major matrix types exist, and the filters of each layer essentially belong to only one type.

The first matrix type is associated with the $512$ filters in layer $13$ only (Fig. 4A). The prototypical matrix consists of nine dominant elements (Fig. 4A, left), and its Boolean version is obtained by clipping its elements using a threshold equals $0.3$ (Fig. 4A, middle). The permutation of this clipped matrix results in a $3 \times 3$ diagonal block matrix with unity elements (Fig. 4A, right). This rearrangement indicates that the subset output labels, e.g. $(1, 5, 8)$, is favored by a filter. Moreover, the observation of solely dominant output fields strongly indicates that the input label belongs to the filter's subset, e.g. $(1, 5, 8)$. Each of the $512$ filters belonging to layer $13$ consists of only one preferred subset of labels, a cluster, which generally consists of $3$ labels and much less often $2$ or $4$ labels (Fig. 5A). These quantitative results are only slightly dependent on the threshold in the range $[0.3, 0.6]$, and similarly for the other two matrix types (Fig. 4 and Supplementary Information).

The second matrix type is associated with the filters of layer $10$, where the mutual correlation between the input label and average output fields is more structured (Fig. 4B, left). The clipped Boolean matrix consists of more elements above-threshold (Fig. 4B, middle) than the first matrix type. Although the matrix seems to be unstructured, a permutation procedure of labels (Supplementary Information) reveals several diagonal block matrices, e.g. two of size $3 \times 3$ and two of size $1 \times 1$ (Fig. 4B, right). The remaining unit elements outside of the diagonal block matrices, clusters, are identified as filter noise (6 yellow units in Fig. 4B, right). Each filter in the $10^{th}$ layer typically consists of several diagonal block matrices with additional noise elements, where the number of clusters decreases as a function of their size and practically vanish above size $3$ (Fig. 5A).

The third matrix type is associated with the filters in layers $2$ and $4$ (Fig. 4C). Each matrix is characterized by a few columns, where almost all their elements are above-threshold, for example, a matrix with two such columns (left) and its clipped Boolean version (middle). Permutation of the clipped matrix results in block diagonal matrices with additional substantial noise (yellow) associated with their columns (right). Layer 7 filters represent a mixture of the second and third matrix types.

The statistical results obtained from four trained VGG-16 samples indicated the following key trends (Fig. 5A). First, each filter in layer $13$ consists of only one diagonal cluster with an average size close to $3$ (Fig. 5A). This type of filter structure indicates with high probability that for an event with only three dominant output fields on the labels constituting the cluster, the input label is one of the three labels. This type of mutual information between the input label and average output fields is enhanced with decreasing the cluster size. It is zero for a cluster of size 10 and is maximized for a unit size. Clearly, the ideal noiseless cluster structure consists of 10 diagonal clusters of size 1; however, this is unattainable because the complex classification task is not linearly separable.

The second trend is that the average noise per filter decreases with layers toward the output, from $\sim 20$ in layer 2 to 1.5 in layer 13 (Fig. 5A). This almost noiseless input environment for the output layer is a necessary condition for achieving high SRs; however, it is not sufficient. For instance, assume 512 noiseless filters with diagonal clusters of size 3 that are composed of labels $[0 - 6]$. Because the clusters do not include labels $[7 - 9]$, low SRs can be expected.

The third trend is a global tendency for equalization among all labels comprising clusters of layer filters, as measured by the number of appearances of each label (Fig. 5B). For layer 13, the average cluster size was close to 3 and there, one cluster exists per filter (Fig. 5A). Hence, the expected number of clusters including a given label can be approximated by $512 \cdot 3 : 10 \sim 150$. The results of a typical sample confirm that the number of occurrence of each label in clusters of layer 13 slightly fluctuates around their average values, $\sim 150$ (Fig. 5B, left). Similarly, for layer 7 with 256 filters, 1.3 clusters per filter, and an average cluster size of 1.66 (Fig. 5A), the average number of occurrence per filter is $256 \cdot 1.3 \cdot 1.66 : 10 \sim 55$ (Fig. 5B), and the appearance of each label slightly fluctuates around it.

Silencing a large fraction of filters, which include a specific label, impairs equalization among labels and substantially decreases the SRs.

**The mechanism underlying successful DL**

By definition, a diagonal filter cluster generates significant output fields on its internal labels (Fig. 4) with additional average negative output fields on the remaining labels (Fig. 5C, left). Moreover, for a noiseless filter, the significant output fields are solely on the cluster labels (Fig. 6A), which is a close realization of layer 13, where also the noise level is minimal (Figs. 4A and 5A). Similarly, when the input label differed from the diagonal cluster labels, its 10 output fields were all, on average, two orders of magnitude smaller than the output fields for labels belonging to the cluster (Figs. 5C, right, and 6B).

In layer 13, each label appeared, on average, in ~150 clusters among the 512 (Fig. 5B), where there was one cluster per filter (Fig. 5A). Hence, an input label induced an output field equals 150 times the average label's output field belonging to a cluster, which, for simplicity, was defined as a unit field (Fig. 6C). This field amplitude represents the signal for selecting the correct output label (Fig. 6C) and must be compared with the noise for selecting a different output label.

Two types of sub-threshold noise exist: one associated with the abovementioned 150 filters and the second generated by the remaining $512 - 150 = 362$ filters. Assume for simplicity, all clusters in layer 13 are of size $3 \times 3$, which is close to their average size, 2.88 (Fig. 5A). Hence, in the 150 filters, each of the nine labels, excluding the input label, appears on average 33 times $(2 \cdot 150 \div 9)$, and generates an output field with an amplitude of ~33. This relatively small field, in comparison to 150 output fields on the input label, decreases even further because, in $150 - 33$ such filters, average negative fields exist on all nine labels (Figs. 5C, left, and 6A). These summed output fields $(< 33)$ represent the first type of noise (Fig. 6C). This is much smaller than the ~150 signal output field on the correct label, resulting in a signal-to-noise ratio (SNR) $> 5$ (Fig. 6C, right). The second type of noise stems from the remaining 362 filters, whose clusters do not include the input label and are expected to induce, on average, a small negative output field (Figs. 5C, right, and 6B).

The prerequisite for the effectiveness of the proposed mechanism underlying the DL is almost noiseless filters, which is realized by decreasing the noise of the layer toward the last CL (Fig. 5A). This results in high SNRs of the output fields, such that the SRs should approach unity, as opposed to ~0.94 (Fig. 2A). However, one must consider that the presented argument is based on the correlation between the input label and averaged output fields, where the fields for a particular input can significantly differ from their average values.

**Examination of VGG-6**

The VGG-6 architecture consists of six layers, similar to VGG-16, but with the following two modifications: five CLs exist with a max-pooling operation after each, and the architecture terminates with only one FC layer (Fig. 7A). The estimation of each layer SR was performed according to the discussed methodology (Fig. 1), resulting in increased SRs with layers up to $0.92$ at layer 5 (Fig. 7B).

The three trends obtained for VGG-16 also characterized VGG-6. In the $5^{th}$ layer each filter consisted of only one cluster, and the noise decreased with increasing layers toward the output (Fig. 7C). In addition, the number of occurrences of each label in the clusters belonging to a layer was similar (not shown).

The decreased average SR of VGG-6 ($0.92$) compared to that of VGG-16 ($0.94$) is mainly attributed to the increase in the average cluster noise (Fig. 7C). The noise in the $5^{th}$ layer of VGG-6 was $3.14$ (Fig. 7C), whereas that for layer 13 of VGG-16 was only $1.5$ (Fig. 5A). This trend dominates the opposite trend, where the average cluster size decreases from $2.88$ in VGG-16 to $2.6$ in VGG-6 (Fig. 7C).

**Examination of AVGG-16**

The Advanced VGG-16 (AVGG-16) architecture consists of 16 layers, but with only two pooling operations: $4 \times 4$ average pooling after the $7^{th}$ CL, and $2 \times 2$ max-pooling after the $13^{th}$ CL (Fig. 8A). AVGG-16 yields an enhanced SR of 0.955 when performing pooling decisions adjacent to the output layer[26].

The improvement of the average SR of AVGG-16 by $\sim 0.015$ compared to that of VGG-16 can be attributed to the following two trends. The cluster noise level, averaged over three trained samples, was equal to $\sim 1.1$, in comparison to $1.5$ for VGG-16 (Fig. 5A). In addition, the average cluster size decreased from $2.88$ for VGG-16 (Fig. 5A) to $2.5$ for AVGG-16 (Fig. 8A). These two trends indicate that the dominant label output fields pinpoint the input label with a higher probability. The decreased noise level focuses on the possible input labels for the cluster labels and a smaller cluster size further reduces the possible options for the input label.

**DISCUSSION**

A method of quantifying the SRs of each CL in a deep architecture is presented. In the first stage, the entire deep architecture is trained to maximize the SRs. In the second stage, the weights of the first $m$ trained layers are held unchanged, and their outputs are FC to the output layer (Fig. 1B-C). The output of the first $m$ layers represents the preprocessing of an input using a partially deep architecture, and the FC layer is trained to maximize the SR, which is a relatively simple computational task. The test set results indicate that the SRs progressively increase with the number of layers toward the output (Fig. 2A).

In the third stage, the trained weights of the FC layer are used to quantify the functionality of each filter belonging to its input layer (Fig. 3A). The single-filter performance is calculated where all weights of the FC layer are silenced, except for the specific weights that emerge from the single filter. Test inputs are now presented and preprocessed by the first $m$ layers but influence the outputs only through the small aperture of one filter. These three stages constitute the algorithmic aspects of the proposed method.

The results indicate that each filter essentially selects a single output label independent of the input label (Fig. 3B), which seems to prevent high SRs. One conclusion could be that the examination of a single filter functionality cannot reveal the mechanism underlying successful DL. Nevertheless, the $10 \times 10$ correlation matrix, representing the input and average-label output fields, reveals an interesting structure. Each filter favors a small subset of input-output labels according to their relatively high output fields (Fig. 4), a feature that significantly increases the knowledge of the possible input label. The emergence of solely high output fields on a small cluster of output units, represented by a diagonal block of the permute matrix (Fig. 4), strongly indicates that the input label belongs to this cluster. However, this characteristic significantly diminishes with filter noise, represented by additional high-output fields outside of the cluster labels. Nevertheless, the results indicate that filter noise progressively decreases toward the final CL (Fig. 5A).

A critical aspect of DL, particularly in the $13^{th}$ CL, is that each input label appears almost equally in all clusters comprising the filters of a given layer (Fig. 5B). In the case of $K$ filters in a layer and an average cluster size of $3$, each label appears in $3 \cdot K : 10 = 0.3 \cdot K$ clusters. This finite fraction of clusters generates a significant output signal on the correct label, where $SNR \sim 5$ (Fig. 6C); however, it has to compete with the following two types of sub-threshold noise. The first type is noise induced by these $0.3 \cdot K$ filters on labels outside their clusters, and the second type is attributed to the generated fields on the correct label from the remaining $0.7 \cdot K$ filters. The amplitudes of both types of noise are much smaller than those of the signal and could even be negative (Fig. 5C). A positive SNR significantly greater than unity is the underlying phenomenon that enables successful DL. It provides a quantitative explanation for the superiority of VGG-16 SRs over VGG6 (Fig. 7) and AVGG-16 SRs (Fig. 8) over VGG-16.

One must bear in mind that the functionality of each filter is estimated based on its averaged output fields over the entire test set. For a particular input, the output fields can deviate significantly from their average values; however, using a large number of filters, for example, $512$ (Fig. 1), typically compensates for these types of fluctuations, and high SRs can be achieved. Nevertheless, randomly removing $\sim150$ filters out of $512$ of the trained VGG-16 layer $13$, which was FC to the output (Fig. 1B), did not affect the SRs, which remained at $\sim0.94$. In addition, training all layers of such a VGG-14 architecture[26], where the last layer consisted of $512 - 150 \sim 350$ filters only, did not affect the SRs as well. This dilution does not violate the necessary conditions for a large SNR, in which each label must appear almost equally in the clusters belonging to the filter of the layer (Fig. 5B). In VGG-16, the average cluster size in layer $13$ was $\sim3$ (Fig. 5A), and the number of different triplets of labels is $10 \cdot 9 \cdot 8 : 6 = 120$, which could be occupied almost equally with $350$ filters only. Similarly, diluting the number of filters in layers $11$ through $13$, to ~$350$ filters, exhibited little to no effect on SRs. Decreasing the number of filters further toward $120$ only slightly affected the SRs. For advanced architectures, where the cluster size may be reduced to ~$2$, the number of different pairs is only $45$, and one might expect the SRs not to be affected by an even smaller number of filters. Indeed, for AVGG-16, the average cluster size in layer $13$ was $2.5$ (Fig. 8), and the SRs remained practically the same, with $100$ filters instead of $512$. Hence, understanding the mechanism underlying DL can direct the methods for simplifying architectures without affecting their SRs. Maintaining SRs with simplified architectures could reduce the computational complexity and test latency (Fig. 2), where the same SRs were achieved with fewer CLs.

A common trend of all the examined deep architectures was that, in the last CL, each filter consisted of only one cluster. Assume for simplicity that all clusters are of size 3 in the last CL. This trend is questionable because it could be better for DL to form a combination of cluster structures, such as two $3 \times 3$ clusters or three $2 \times 2$ clusters per filter. These types of structured filters do not change the above-mentioned SNR argument. However, the interference between a larger subset of preferred output labels increases the noise level. For instance, in VGG-16 layer 10, each filter consisted of a few clusters, but the noise level increased to 3.8, in comparison to 1.5 in layer 13 where each filter consisted of only one cluster (Fig. 5A). It appears difficult for DL to reduce noise while simultaneously forming structured filters composed of several clusters each. Although the noise of layer 10 is higher than layer 13's, it yields similar SRs which can be attributed to its smaller average cluster size (Fig. 5A).

Statistical features were presented only for layers terminating with MP operators, layers $2, 4, 7, 10,$ and $13$ of VGG-16 (Fig. 2). For the other layers, the statistical features indicated that the noise was not monotonic and was typically significantly increased in the layers subsequent to the MP, for example, layers $8$ and $11$. This behavior is attributed to the relaxation process for better rearrangement of the filter clusters along a block of several consecutive CLs.

The training set differed from the test set, where the SR increased rapidly toward 1, even for an intermediate layer trained with an FC layer to the output layer (Fig. 1). For instance, layer $7$ in VGG-16 achieved an SR of ~0.999 on the training set. Nevertheless, this characteristic is the only apparent feature that distinguishes the training and test sets. Other features of the training set, such as the distribution of cluster sizes and noise levels, were almost identical for layer $13$, indicating that a zero training error could not be simply deduced from the statistical features. Nevertheless, the identical features of both the training and test sets may direct a procedure to eliminate or fix selected filters during training to enhance the SRs.

Understanding a physical phenomenon is the primary scientific goal that serves as a source for revealing new discoveries and practical advantages.

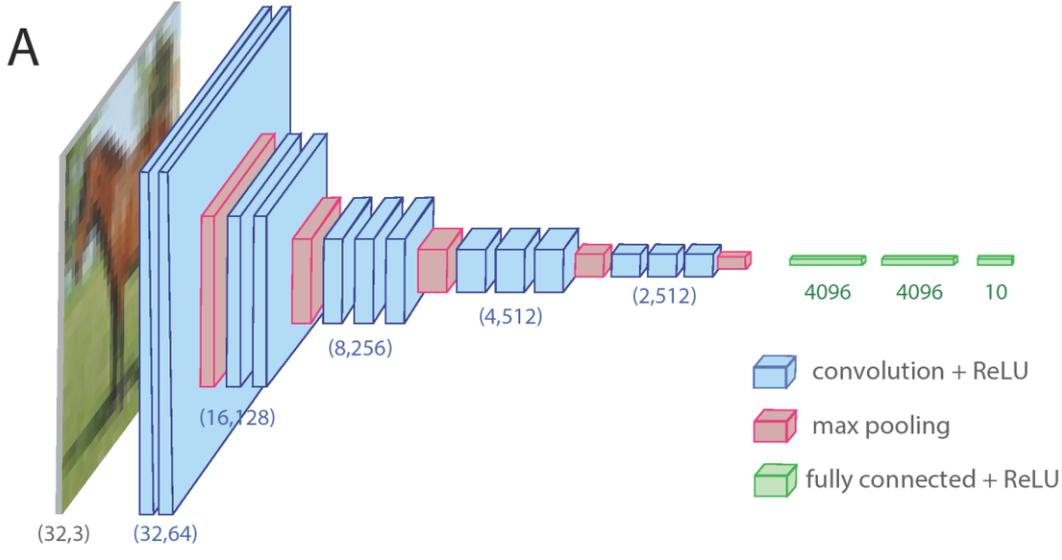

A

(32,3)  (32,64)  (16,128)  (8,256)  (4,512)  (2,512)  4096  4096  10

convolution + ReLU

max pooling

fully connected + ReLU

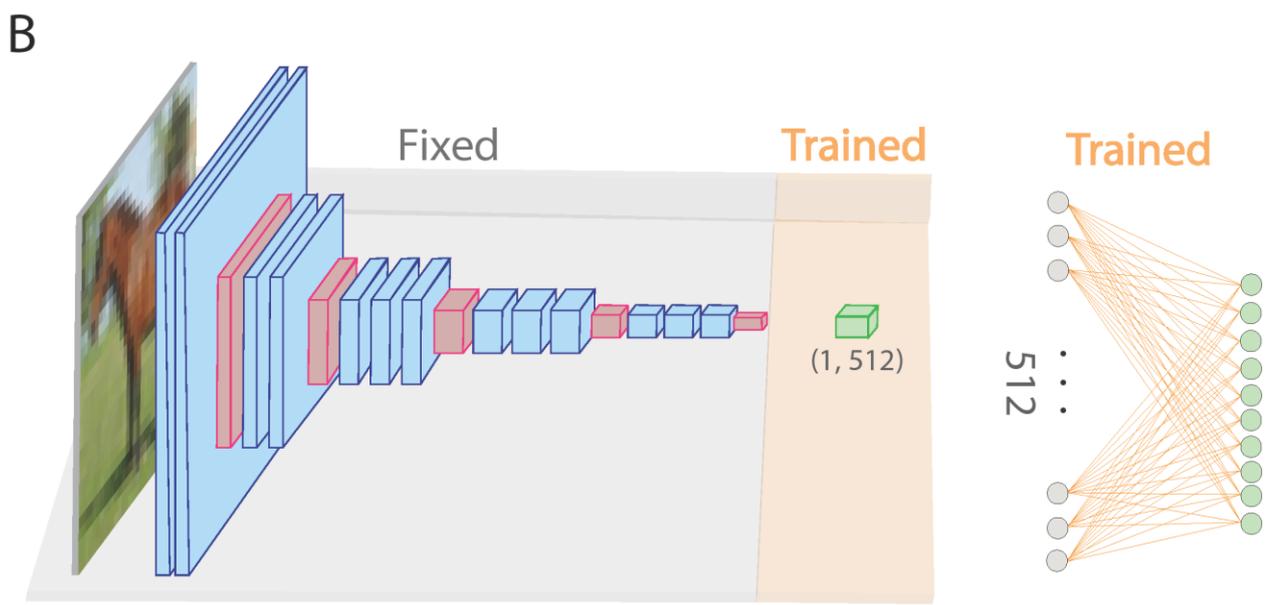

B

Fixed    Trained    Trained

(1, 512)

512

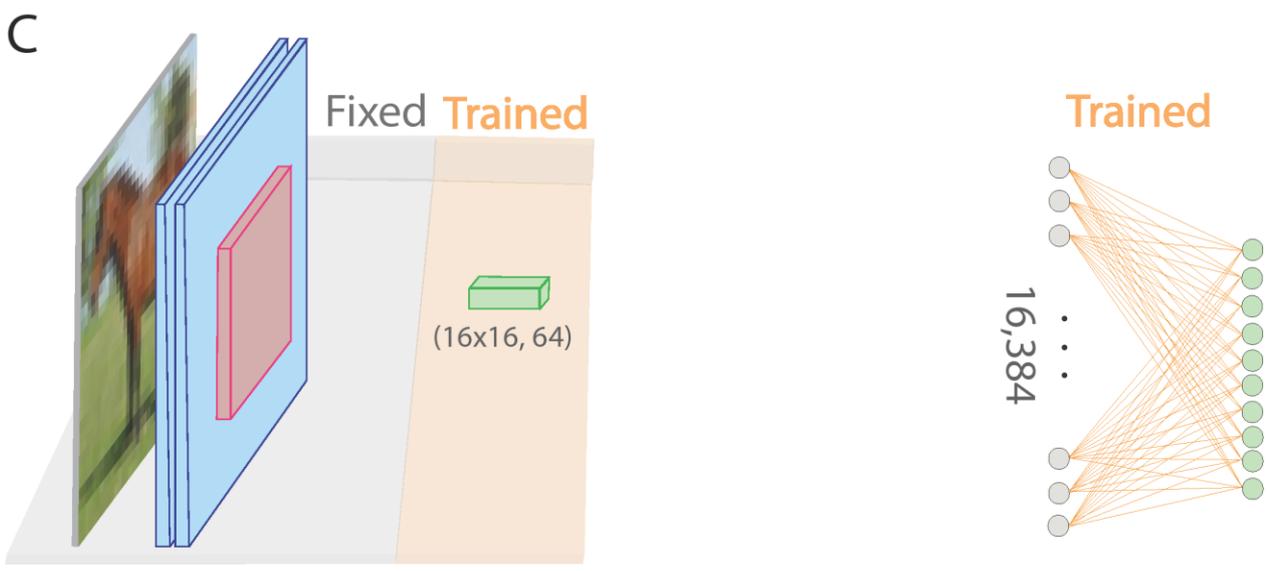

C

Fixed    Trained    Trained

(16x16, 64)

16,384

**Figure 1. Methodology for measuring progressive layer SRs, exemplified on VGG-16 architecture. (A)** VGG-16 architecture for classification of the CIFAR-10 database, consisting of 13 convolutional layers (CLs) with ReLU activation function, 5 max-pooling (MP) operators, 3 fully connected (FC) layers, and 10 output units. **(B)** First 13 CLs of VGG-16 with fixed weights, where their $N(13) = 512$ outputs after the MP (left) are connected to the 10 outputs with trained weights (right, orange). **(C)** Similar to **B**, where $N(2) = 16,384$ outputs after MP of the first two fixed CLs (left) are connected with trained weights to the output units (right, orange).

A

| Layer | No. filters | Filter's outputs | FC inputs | SR | Std |
|-------|-------------|------------------|-----------|-------|-------|
| 13 | 512 | 1x1 | 512 | 0.94 | 0.002 |
| 10 | 512 | 2x2 | 2,048 | 0.94 | 0.001 |
| 7 | 256 | 4x4 | 4,096 | 0.931 | 0.001 |
| 4 | 128 | 8x8 | 8,192 | 0.852 | 0.003 |
| 2 | 64 | 16x16 | 16,384 | 0.725 | 0.004 |

B

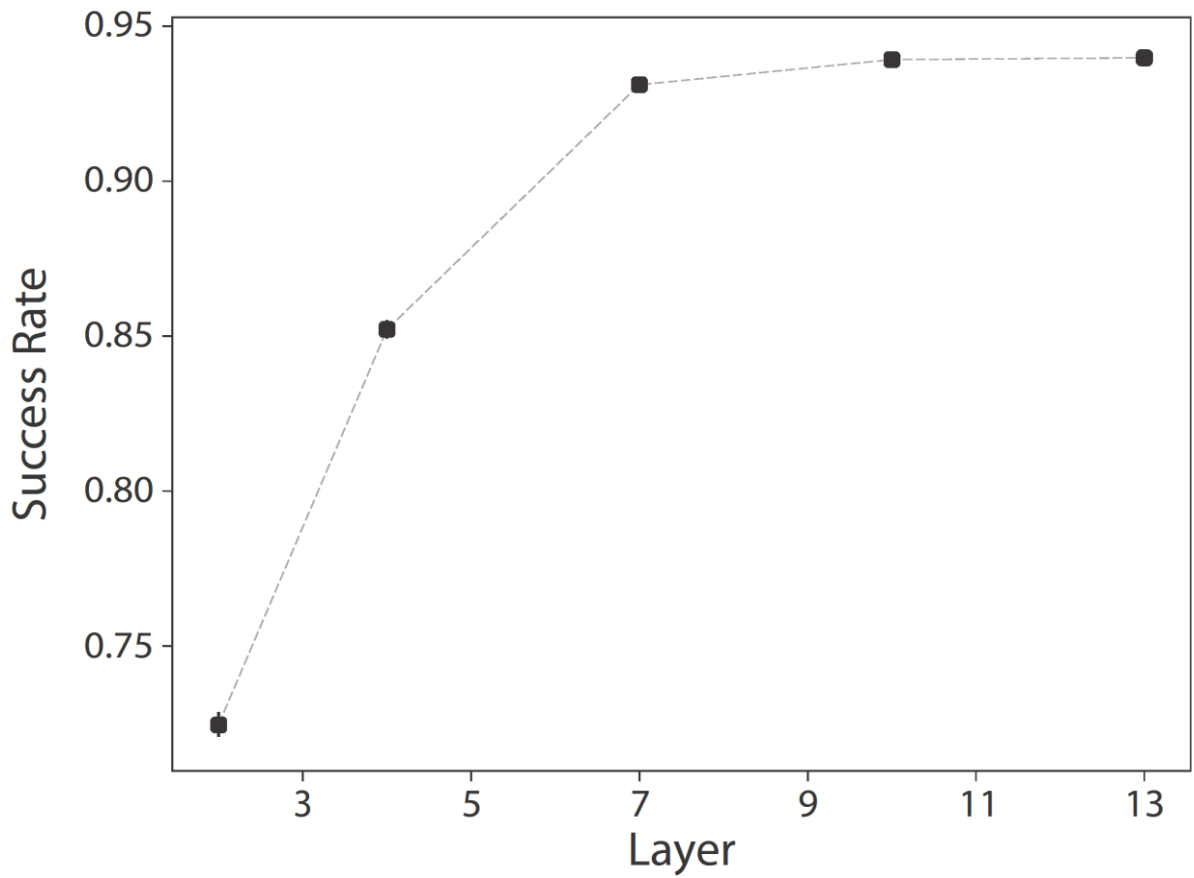

**Figure 2. Progressive layer SRs for VGG-16. (A)** Maximized SRs and their standard deviations (Stds) for fixed weights of the first $m = 2, 4, 7, 10,$ and 13 VGG-16 layers and their trained FC layer with $N(m)$ inputs, after MP, to the output layer (Supplementary Information). **(B)** SRs and their Stds as a function of the number of measured layers.

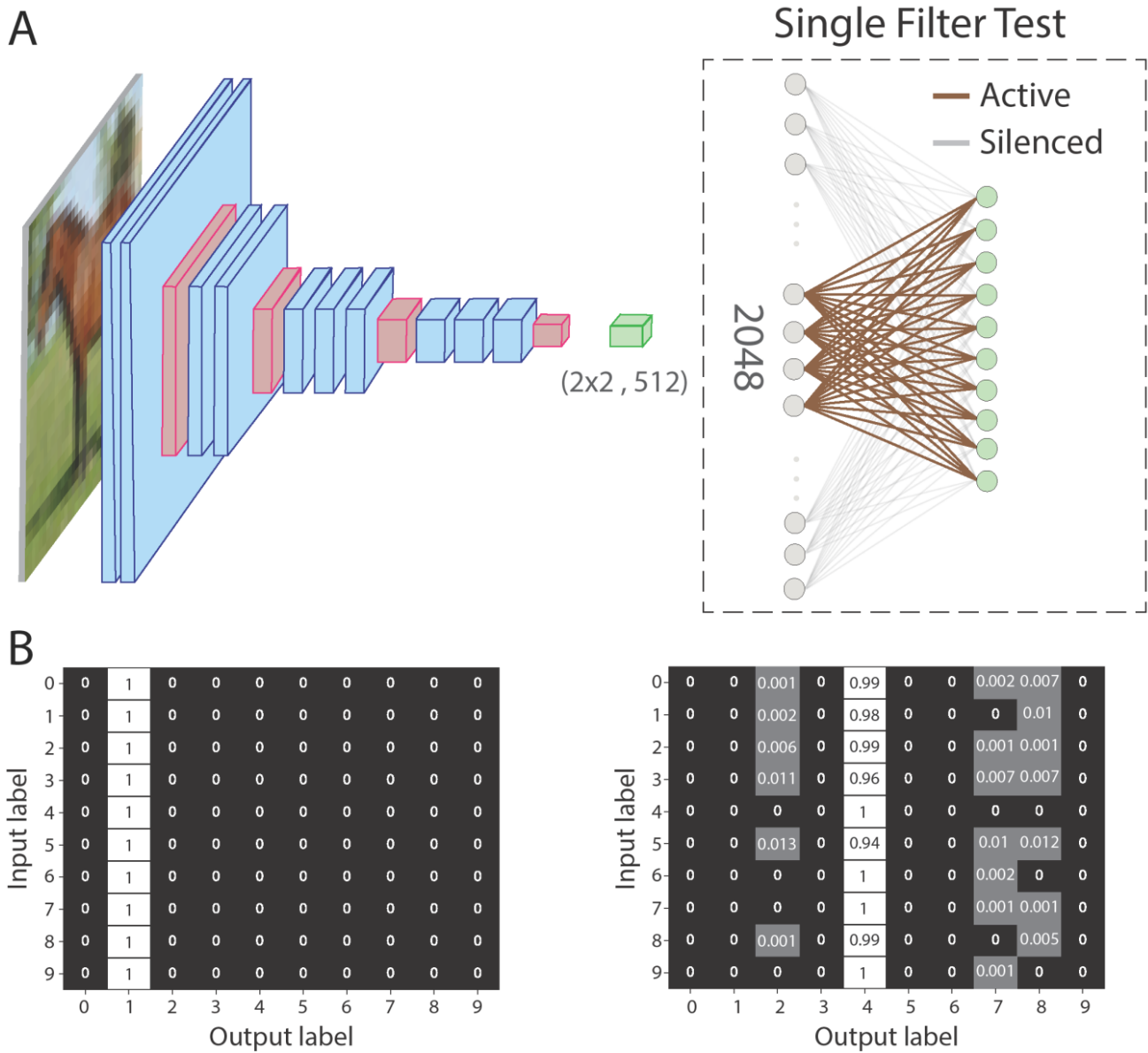

**Figure 3. Single filter SR. (A)** Example where the first $m = 10$ CLs of VGG-16 are concatenated with a trained FC layer consisting of $2048$ inputs, $4$ for each of the $512$ filters. All weights of the FC layer are silenced (light gray) except for the emerging weights from the selected filter (brown). **(B)** Representative filter in layer 13 where the mutual input-output label probability indicates the selection of a single output label independent of the input label (left). Similarly, a filter in layer 4 selecting mainly two output labels (right). Test inputs with zero fields on all outputs are excluded (Supplementary Information).

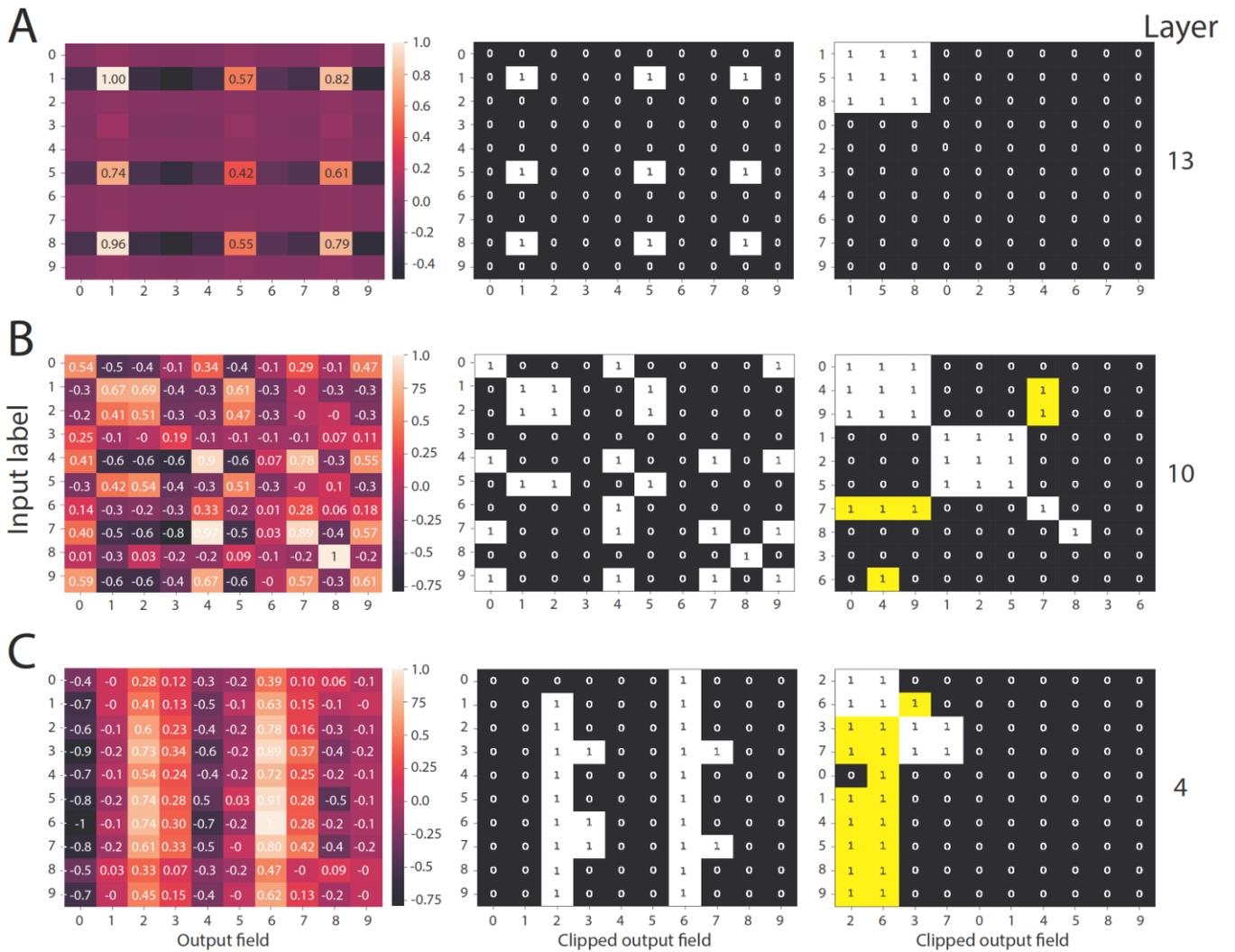

**Figure 4. Single filter performance. (A)** The matrix element $(i, j)$ of a filter belonging to layer 13 stands for the averaged fields generated by label $i$ test inputs on an output $j$, where the matrix elements are normalized by their maximal element (left). The Boolean clipped matrix following a given threshold (middle). Permutations of the clipped matrix labels resulting in a block diagonal (right). **(B)** Similar to **A** for a filter in layer 10, where permutation of the clipped matrix results in several diagonal blocks with additional noise elements (yellow). **(C)** Similar to **B** for a filter in layer 4, where the Boolean clipped matrix (middle) of almost all elements of the two columns are above-threshold and noise (yellow) almost completely fills the two columns (right).

**A**

| Layer | Av. Noise | Av. Clusters/filter | Av. Cluster size | Cluster size fraction | | | | | |
|-------|-----------|---------------------|------------------|---|---|---|---|---|---|
| | | | | 1 | 2 | 3 | 4 | 5 | 6 |
| 13 | 1.5 | 1 | 2.88 | 0.004 | 0.19 | 0.67 | 0.12 | 0.002 | 0 |
| 10 | 3.8 | 3.2 | 1.62 | 0.51 | 0.37 | 0.11 | 0.01 | 0.0002 | 0 |
| 7 | 6.4 | 1.3 | 1.66 | 0.50 | 0.37 | 0.11 | 0.02 | 0.0007 | 0 |
| 4 | 18.3 | 1.46 | 2.1 | 0.36 | 0.28 | 0.27 | 0.08 | 0.009 | 0 |
| 2 | 19.6 | 1.36 | 2.24 | 0.333 | 0.30 | 0.26 | 0.1 | 0.017 | 0.005 |

**B**

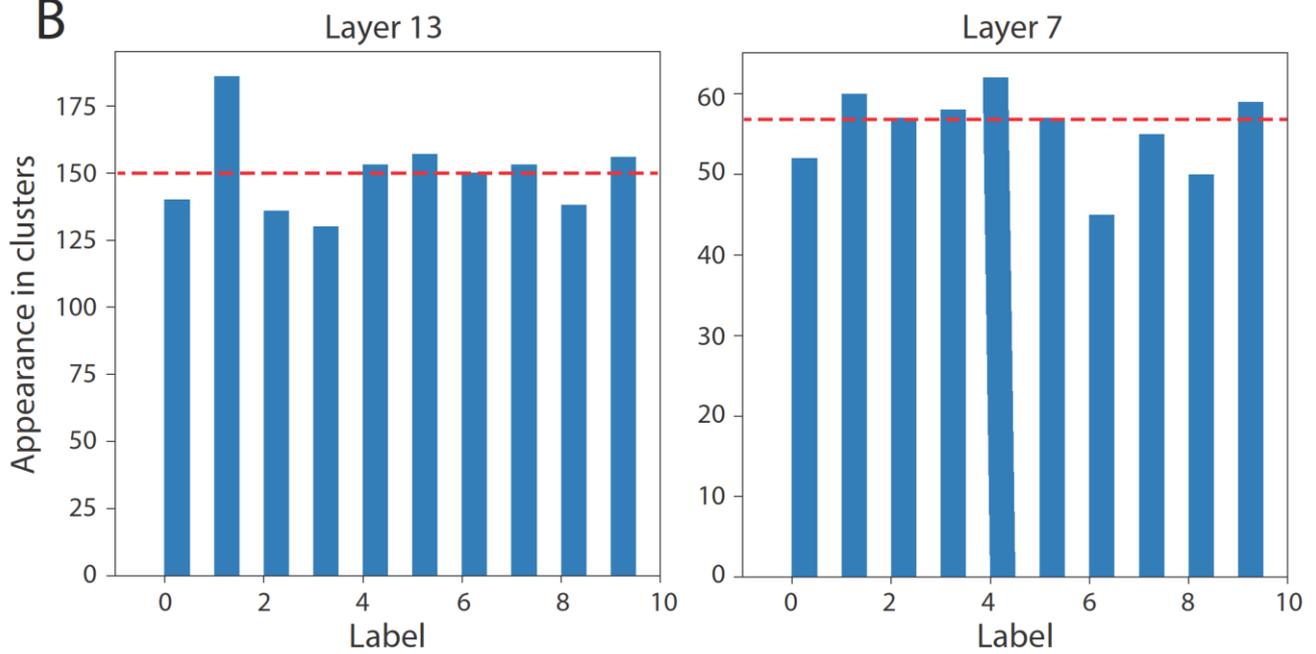

**C**

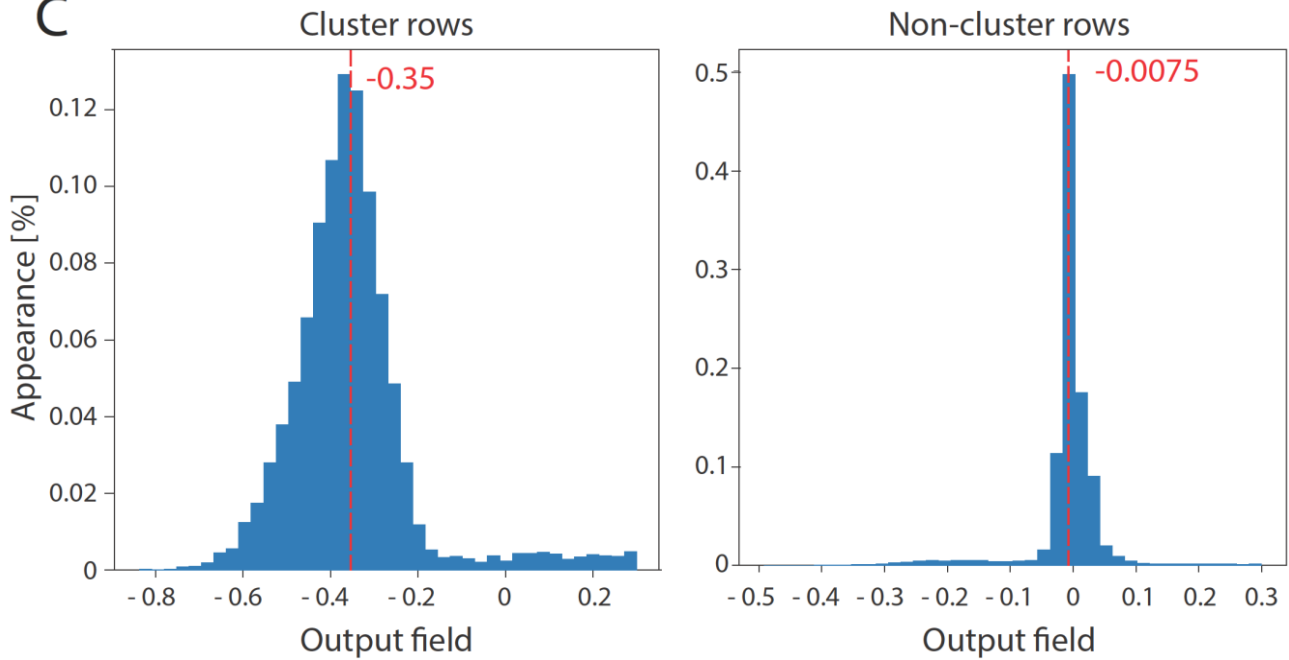

**Figure 5. Statistical features of filters per layer in VGG-16. (A)** Averaged results obtained from four trained VGG-16 samples. **(B)** Appearance number of each label in the diagonal clusters of the 512 filters of layer 13, obtained from a representative sample (left), including the average number (dashed-red horizontal line) and for layer 7 with 256 filters (right). **(C)** Output field histogram and its average value (dashed-red vertical line) of elements belonging to cluster rows of the 512 normalized matrices of layer 13 (e.g. Fig. 4A, left matrix), excluding elements above-threshold (left). Similarly, for non-cluster rows (right).

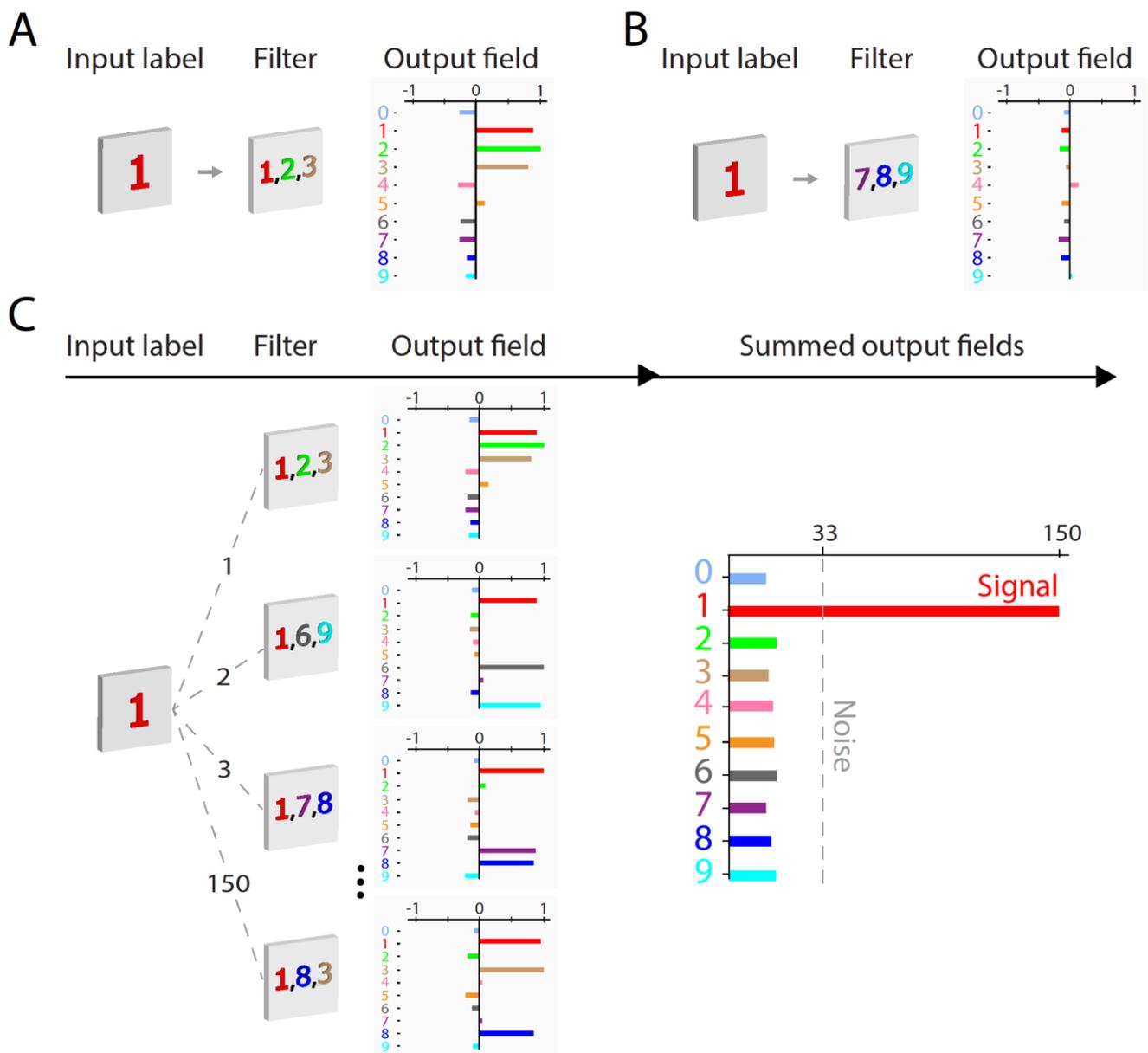

**Figure 6. The mechanism underlying successful DL. (A)** Scheme of an input label 1 to a $3 \times 3$ diagonal filter composed of labels 1, 2, and 3, resulting in output fields ~1 on these three labels, and on average a small negative noise on the rest of the labels (Fig. 5C, left). **(B)** Similar to **A**, where the diagonal filter does not include label 1, resulting in small negative or positive noise on all labels (Fig. 5C, right). **(C)** Scheme of signal-to-noise-ratio (SNR), where label 1 is presented for the 150 filters of layer 13, whose clusters include label 1, and their output fields are similar to those in **A** (left). The summed 150 filter output field results in a signal with label 1 and noise on the other labels, where SNR > 5 (right).

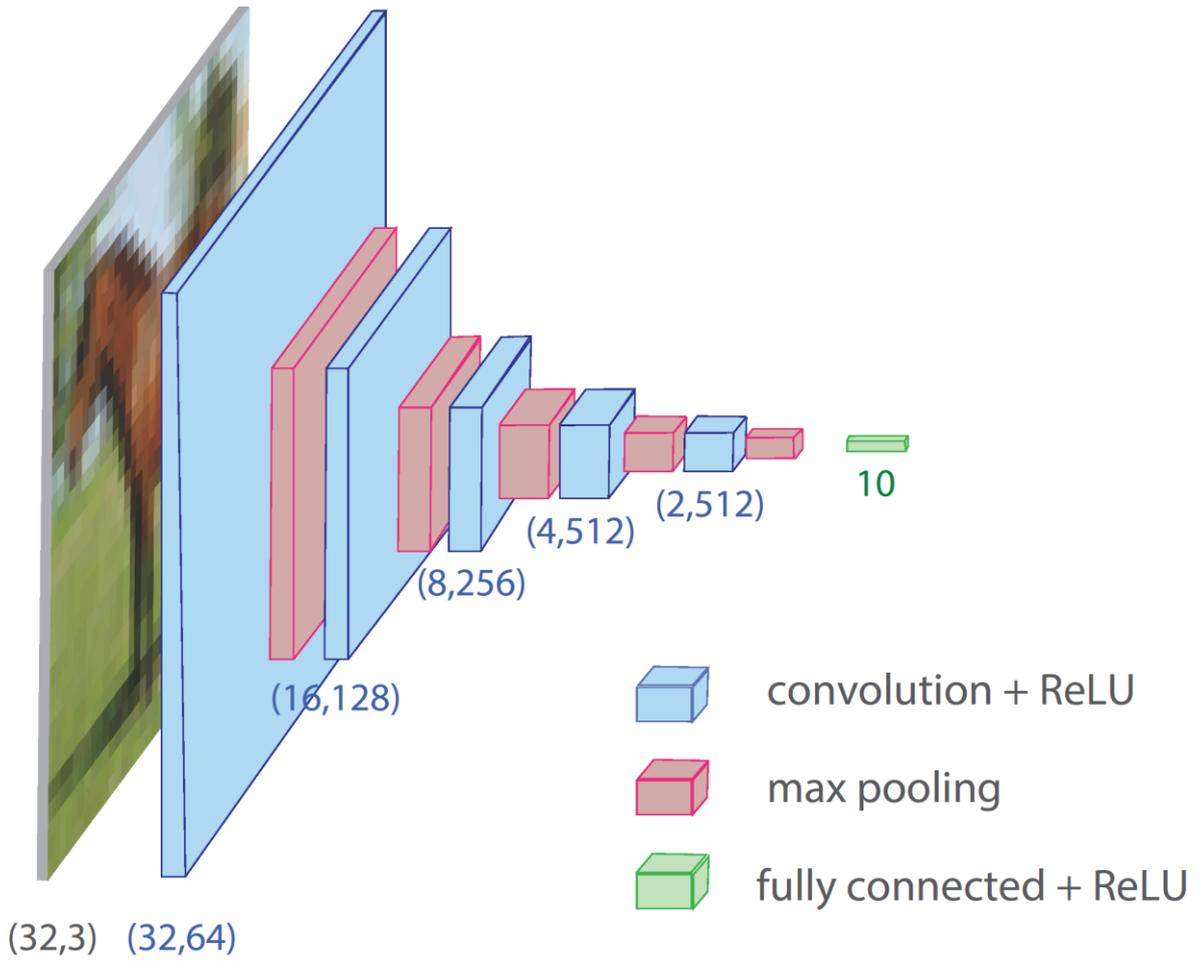

A

(16,128)
(8,256)
(4,512)
(2,512)
10

convolution + ReLU

max pooling

fully connected + ReLU

(32,3)　(32,64)

B

| Layer | No. Filters | Filter's outputs | FC Inputs | SR | Std |
|-------|-------------|------------------|-----------|------|-------|
| 5 | 512 | 1x1 | 512 | 0.92 | 0.001 |
| 4 | 512 | 2x2 | 2,048 | 0.91 | 0.004 |
| 3 | 256 | 4x4 | 4,096 | 0.88 | 0.008 |
| 2 | 128 | 8x8 | 8,192 | 0.8 | 0.008 |
| 1 | 64 | 16x16 | 16,384 | 0.7 | 0.006 |

C

| Layer | Av. Noise | Av. Clusters/filter | Av. Cluster size | Cluster size fraction | | | | | |
|-------|-----------|---------------------|------------------|------|------|------|------|-------|-------|
| | | | | 1 | 2 | 3 | 4 | 5 | 6 |
| 5 | 3.14 | 1 | 2.6 | 0.12 | 0.36 | 0.35 | 0.14 | 0.02 | 0.002 |
| 4 | 5.47 | 1.1 | 1.4 | 0.66 | 0.26 | 0.06 | 0.01 | 0 | 0 |
| 3 | 12.28 | 1.31 | 1.7 | 0.48 | 0.35 | 0.14 | 0.03 | 0.002 | 0 |
| 2 | 18.16 | 1.42 | 2.2 | 0.34 | 0.34 | 0.23 | 0.07 | 0.03 | 0.001 |
| 1 | 18.5 | 1.3 | 2.1 | 0.4 | 0.33 | 0.2 | 0.1 | 0.008 | 0.002 |

**Figure 7. Statistical features of SRs and filters per layer in VGG-6. (A)** VGG-6 architecture for classification of the CIFAR-10 database, consisting of five convolutional layers (CLs) with ReLU activation function, where each FC layer terminates with a $2 \times 2$ max-pooling (MP) operator which finally FC to the 10 output units. **(B)** Maximized SRs and their standard deviations (Stds) for fixed weights for each of the layers and their trained FC sizes. **(C)** Averaged results obtained from five trained VGG-6 samples, indicating decreased averaged layer noise toward the output, where each filter in layer 5 consists of one cluster.

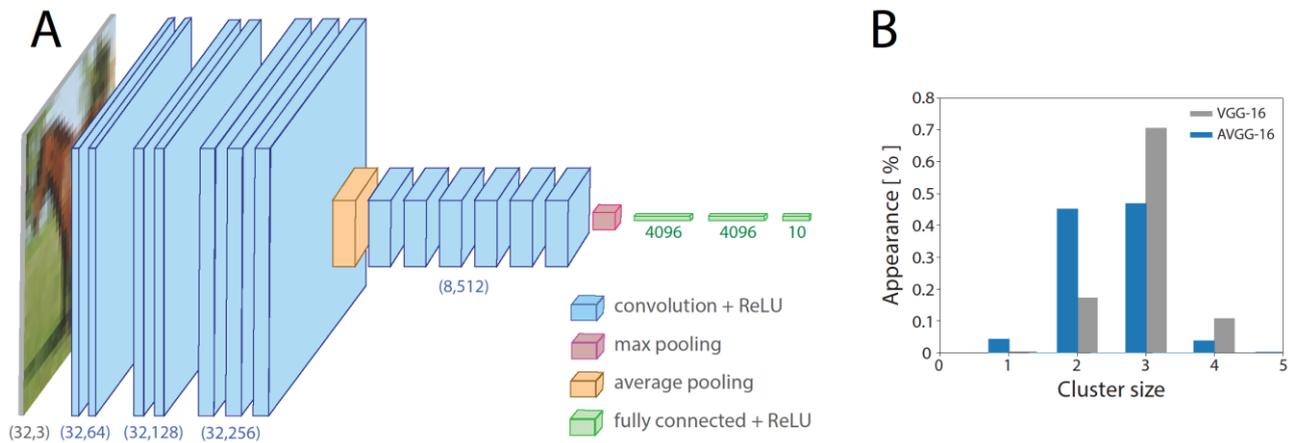

**Figure 8. AVGG-16 architecture and statistical features of filters per layer. (A)** AVGG-16 architecture for classification of the CIFAR-10 database, consisting of 13 convolutional layers (CLs) with ReLU activation function, $4 \times 4$ average-pooling operator after layer 7, $2 \times 2$ max-pooling operator after layer 13, 3 fully connected layers, and 10 output units. **(B)** Representative distribution of cluster sizes for AVGG-16 (blue), obtained from three trained samples, and for VGG-16 (gray), obtained from four trained samples. The average cluster size is 2.5 for AVGG-16 and ~2.9 for VGG-16.

**Data availability**

Source data are provided in this study, including all data supporting the plots, along with other findings of this study.

**Acknowledgments**

I.K. acknowledges the partial financial support from the Israel Science Foundation (grant number 346/22). S.H. acknowledges the support from the Israeli Ministry of Science and Technology.

**Author contributions**

Y.T. led all the simulations with the help of Y.M., O.T. and R.D.G. contributed to some of the simulations. S.H. contributed to some simulations and prepared the figures. R.V. discussed the results and commented on them. I.K. initiated the study, supervised all aspects of the study and wrote the manuscript. All the authors commented on the manuscript.

**Competing interests**

The authors declare no competing interests.

# Supplementary Information

## The mechanism underlying successful deep learning

**Yarden Tzach[1], Yuval Meir[1], Ofek Tevet[1], Ronit D. Gross[1], Shiri Hodassman[1], Roni Vardi[2] and Ido Kanter[1,2*]**

[1]Department of Physics, Bar-Ilan University, Ramat-Gan, 52900, Israel.

[2]Gonda Interdisciplinary Brain Research Center, Bar-Ilan University, Ramat-Gan, 52900, Israel.

[*]Corresponding author email: ido.kanter@biu.ac.il

**Architectures and Training the fully connected layer.** Three different architectures were examined, VGG-16[1], VGG-6[2] and AVGG-16[2]. The architectures were trained to classify the CIFAR-10 dataset using hyper-parameters listed below (see Supplementary Information for Figs. 1, 7 and 8). All three architectures were trained with no biases on the 10 output units. This was done to assure that each filter's effect on the output fields will be exemplified and will not be overshadowed by the much larger biases. Removing the 10 biases of the output layer did not change the architectures' average SRs, in comparison to architectures who were trained with output biases. Furthermore, removing the 10 output biases of systems who were trained with output biases, did not affect the systems' SRs.

The examination process was done by taking each system at designated layers and training a fully connected (FC) layer on the output of that specific layer. The FC layer consisted of 10 outputs representing the labels of the CIFAR-10 dataset and input size which corresponds to the output of the specific layer which was examined. During the training of the FC layer, weights and biases of the system remain fixed. For VGG-16 the input units to the FC layers were selected after the max-pooling operations adjacent to layers 2, 4, 7, 10, 13, for VGG-6 after the max-pooling operations of layers 1, 2, 3, 4, 5 and for AVGG-16 only layer 13 was examined.

For each examined layer $m$, the output of the training set for the $m^{th}$ layer was used as a preprocessed dataset to train the FC layer, using the hyper-parameters $\eta = 0.1$, $\mu = 0.975$, $\alpha = 1e - 5$, with a learning rate scheduler of $q = 0.6$ every 20 epochs for VGG-6 and VGG-16, and for AVGG-16 $\eta = 0.02$, $\mu = 0.995$, $\alpha = 1e - 8$, with a learning rate scheduler of $q = 0.6$ every 20 epochs.

**Data preprocessing.** Each input pixel of an image ($32 \times 32$) from the CIFAR-10 database was divided by the maximal pixel value, 255, multiplied by 2, and subtracted by 1, such that its range was $[-1, 1]$. In all simulations, data augmentation was used, derived from the original images, by random horizontally flipping and translating up to four pixels in each direction.

**Optimization.** The cross-entropy cost function was selected for the classification task and was minimized using the stochastic gradient descent algorithm[4,5]. The maximal accuracy was determined by searching through the hyper-parameters (see below). Cross-validation was confirmed using several validation databases, each consisting of 10,000 random examples from the training set, as in the test set. The averaged results were in the same standard deviation (Std) as the reported average success rates. The Nesterov momentum[3] and L2 regularization method[4] were applied.

**Hyper-parameters.** The hyper-parameters η (learning rate), μ (momentum constant[3]), and α (regularization L2[4]) were optimized for offline learning, using a mini-batch size of 100 inputs. The learning rate decay schedule[5,6] was also optimized such that it was multiplied by the decay factor, q, every $\Delta t$ epochs, and is denoted below as $(q, \Delta t)$.

**Fig. 1.** VGG-16 Hyper-parameters.

VGG-16 was trained using the following hyper-parameters to maximize SRs:

| VGG-16 | | | | |
|---|---|---|---|---|
| Layer | η | μ | α | epochs |
| CLs | 0.028 | 0.975 | 1.5e-3 | 200 |
| FC | 0.028 | 0.975 | 1.5e-3 | 200 |

The decay schedule for the learning rate is defined as follows:

$$(q, \Delta t) = (0.6, 20)$$

Each layer $m$ was FC to the 10 outputs via $N(m) \cdot 10$ weights. The FC layer was trained using the hyper-parameters: $\eta = 0.1$, $\mu = 0.975$, $\alpha = 1e-5$, with a learning rate scheduler of $q = 0.6$ every 20 epochs while the rest of the system's weight values and biases remained fixed.

**Fig. 3.** For each layer $m$, the weights of the FC layer, connecting that layer to the output layer, are silenced except for those emerging from a specific filter, thereby showing that filter's contribution to the total output field. The system's classification decision made for each test input based solely upon the weights emerging from a single filter result in a decision matrix, displayed in Fig. 3B, where each element $(i, j)$ represents how many times an input label $i$ was classified as an output $j$. Inputs that yielded zero fields on all the 10 output units were not taken into account. This entire zero output field occurs due to the ReLU activation function emerging from the previous $m^{th}$ convolutional layer that zeros all outputs that are non-positive for the examined filter. Each row was normalized in accordance with the number of label's $i$ non-zero field outputs. In panel B, left, a filter who picks label 1 regardless of the input is observed, while on the right, a filter with a slightly more spread decision style is observed. Results are rounded to the presented number of digits.

**Fig. 4.** Left column, the 10 output fields obtained solely from the weights emerging from a single filter were summed over all 10,000 inputs of the test set, resulting in a $10 \times 10$ matrix where each element $(i,j)$ represents the summed field of output field $j$ for all test set inputs of label $i$. The matrix was then normalized by dividing by its maximal value, resulting in each matrix having a maximal value of 1. This output fields' matrix is displayed in panels A, B, C for layers 13, 10, 4 appropriately to show the different behavior of the filters as they progress in the layers. In the center column the clipped Boolean output field matrix is displayed, where each element whose value is above a threshold (0.3) is set to 1 and all others are zeroed.

In the right column, the labels' axes are permuted such as all labels belonging to a cluster are grouped together consecutively, thereby displaying the clusters in an adjacent fashion where they are displayed as a diagonal block of elements with value 1. Each cluster is defined as a subset of $n$ indices where for each $i,j \in n$ elements $(i,j)$ have the value of 1. The minimal cluster size can be 1, that is one element on the diagonal or 10, the entire matrix. The elements that are equal to 1 are then colored as white, representing that they belong to a cluster in the filter, while non-cluster elements with the value of 1 are classified as above-threshold noise and are colored yellow (note that noise is defined as above-threshold for all figures except Fig. 5C. and Fig. 6A-B). Panel A, middle column matrix has 1 cluster of size $3 \times 3$ comprised of labels (1,5,8) which is permuted in the right column to be better displayed as a single block in the top left matrix. Panel B's middle matrix's clusters can be better viewed in the right column, the matrix has 4 clusters, two of size $3 \times 3$ (0,4,9) and (1,2,5) and two of size 1 (7) and (8) and the axes are permuted to be of order [0,4,9,1,2,5,7,8,3,6].

The calculation of the clusters was done by running along the diagonal, from index (0,0) to (9,9) where the first $(i,i)$ element to have a value of 1 is initially designated as a cluster of size $1 \times 1$. The next $(j,j)\ where\ j \neq i$ element to have a value of 1 is then checked to see if can complete a cluster with $(i,i)$, if yes, then it is added to the cluster and the next diagonal element to have a value of 1 is checked if completes a cluster with $i$ and $j$, if yes it is appended to the cluster, if not the system continues to the next diagonal element. This process is repeated for all value 1 elements in the diagonal as long as there are elements who do not belong to a cluster. Note that this process is not uniquely defined, the order by which the indices are iterated can change the outcome of the clustering process, such as a filter with two clusters of sizes $3 \times 3$ and $1 \times 1$ retrieved by iterating from 0 to 9 can yield in certain very rare scenarios, two clusters of size $2 \times 2$. While possibly alternating the results of a single filter, the overall obtained averaged results remain the same when performing the cluster creation while iterating in a reversed order, since those scenarios are very rare and occur in a negligible number of filters.

The noise is calculated for each filter as the elements with value 1 who do not belong to any cluster. They can be seen in color yellow in the right column.

**Fig. 5.** Panel A, the *Av. Noise* represents the average noise per filter in that specific layer, where the noise counts the 1's who do not belong to any cluster in the clipped Boolean Matrix (Fig. 4). The *Av. Clusters/Filter* is the average number of clusters per filter, this represents the sole number of clusters regardless of their size that are exhibited in average in each filter. The *Av. Cluster size* is the average size of the clusters for each layer, a more thorough representation of that value is displayed on its right through the cluster size fraction, where for each layer the percentage of appearance of that cluster size is displayed. The average size is then calculated by summing the cluster size times the probability of its appearance (the fraction). In Panel B, the number of appearances of each label in all clusters belonging to layer 13 is portrayed. The average is around 150, since the average size of each cluster is $\sim 3$ and there are 512 filters, assuming homogeneity will result in 150 clusters for each label since $3 \cdot 512 : 10 \approx 150$.

Panel C left, displays a histogram of the values in the $512$ $10 \times 10$ field matrices of layer 13 (Fig. 4A, left) of elements in rows whose indices belong to clusters, excluding above-threshold elements. Right, a histogram of the values in the $512$ $10 \times 10$ field matrices of layer 13 (Fig. 4A, left) of the sub-threshold elements outside of rows whose indices belong to a cluster. Note that Fig. 5C includes only sub-threshold elements, and thus did not account for values who are above the threshold.

**Fig. 7. VGG-6 Hyper-parameters**

VGG-6 was trained using the following hyper-parameters to maximize SRs:

| VGG-6 | | | | |
|---|---|---|---|---|
| Layer | η | μ | α | epochs |
| CLs | 0.0145 | 0.97 | 1e-3 | 200 |
| FC layer | 0.002 | 0.975 | 1.2e-3 | 200 |

The decay schedule for the learning rate is defined as follows:

For CLs:

$$(q, \Delta t) = \begin{cases} (0.65, 20) & \text{epoch} \leq 140 \\ (0.55, 20) & \text{epoch} > 140 \end{cases}$$

For the FC layer:

$$(q, \Delta t) = \begin{cases} (0.65, 20) & \text{epoch} \leq 140 \\ (0.5, 20) & \text{epoch} > 140 \end{cases}$$

Each layer $m$ was FC to the 10 outputs using size $N(m) \cdot 10$ weights. The FC was trained using the hyper-parameters: $\eta = 0.1$, $\mu = 0.975$, $\alpha = 1e - 5$, with a learning rate scheduler of $q = 0.6$ every 20 epochs while the rest of the system's weight values and biases remained fixed.

**Fig. 8. AVGG-16 Hyper-parameters**

AVGG-16 was trained using the following hyper-parameters to maximize SRs:

| AVGG-16 | | | | |
|---|---|---|---|---|
| Layer | η | μ | α | epochs |
| CLs | 0.00721 | 0.98 | 1.15e-3 | 280 |
| FC layers | 0.0045 | 0.982 | 1.35e-3 | 280 |

The decay schedule for the learning rate is defined as follows:

For CLs:

$$(q, \Delta t) = \begin{cases} (0.65, 20) & \text{epoch} \leq 140 \\ (0.55, 20) & \text{epoch} > 140 \end{cases}$$

For FC layers, with 10 epochs out of phase:

$$(q, \Delta t) = \begin{cases} (0.65, 20) & \text{epoch} \leq 150 \\ (0.55, 20) & \text{epoch} > 150 \end{cases}$$

Each layer $m$ was FC to the 10 outputs using $N(m) \cdot 10$ weights. The FC layer was trained using the hyper-parameters: $\eta = 0.02$, $\mu = 0.995$, $\alpha = 1e - 8$, with a learning rate scheduler of $q = 0.6$ every 20 epochs while the rest of the system's weight values and biases remained fixed.

**Robustness to different threshold values.** Alternating the threshold between the values of $[0.3, 0.6]$ for VGG-16 did not change the behavior of the quantifiable features. For all thresholds within this range, layer 13 yielded the lowest noise level $\sim 1.5$ and each filter in layer 13 is still comprised of a single cluster per filter. Results demonstrate the robustness of the quantifiable features to different parameters such as the threshold. Furthermore, the monotonous decrease of the noise with layers was still exemplified for all threshold within this range.

**Statistics.** Statistics for VGG-6 were obtained using 5 samples, for VGG-16 using 4 samples and for AVGG-16 using 3 samples.

**Hardware and software**. We used Google Colab Pro and its available GPUs. We used Pytorch for all the programming processes.